\documentclass{article}

\usepackage[preprint]{neurips_2025}

\usepackage[utf8]{inputenc} 
\usepackage[T1]{fontenc}    
\usepackage{url}            
\usepackage{booktabs}       
\usepackage{amsfonts}       
\usepackage{nicefrac}       
\usepackage{microtype}      
\usepackage{xcolor}         
\usepackage[pagebackref,breaklinks,colorlinks,allcolors=iccvblue]{hyperref}
\usepackage{multirow}
\usepackage{url}
\usepackage{graphicx}
\usepackage{xspace}
\usepackage{colortbl}
\usepackage{wrapfig}
\usepackage{caption}

\usepackage{amsmath,amsfonts,bm}

\def\eqref#1{equation~\ref{#1}}

\def\1{\bm{1}}

\DeclareMathAlphabet{\mathsfit}{\encodingdefault}{\sfdefault}{m}{sl}
\SetMathAlphabet{\mathsfit}{bold}{\encodingdefault}{\sfdefault}{bx}{n}

\definecolor{iccvblue}{rgb}{0.21,0.49,0.74}
\newcommand{\Mname}{VC$^2$L\xspace}

\usepackage{etoolbox, subcaption}

\newdimen\abovecrulesep
\newdimen\belowcrulesep
\makeatletter
\patchcmd{\@@@cmidrule}{\aboverulesep}{\abovecrulesep}{}{}
\patchcmd{\@xcmidrule}{\belowrulesep}{\belowcrulesep}{}{}

\definecolor{demphcolor}{RGB}{144, 144, 144}
\definecolor{mygray}{gray}{0.4}
\definecolor{lightgray}{rgb}{0.9, 0.9, 0.9}

\newlength\savewidth

\newcommand{\tablestyle}[2]{\setlength{\tabcolsep}{#1}\renewcommand{\arraystretch}{#2}\centering\footnotesize}
\makeatletter\renewcommand\paragraph{\@startsection{paragraph}{4}{\z@}{0em\@plus1ex\@minus.2ex}{0em}{\normalfont\normalsize\bfseries}}
\makeatother

\newcolumntype{C}[1]{>{\centering\arraybackslash}p{#1}}
\newcolumntype{R}[1]{>{\raggedleft\arraybackslash}p{#1}}
\newcolumntype{L}[1]{>{\raggedright\arraybackslash}p{#1}}

\usepackage{etoolbox}
\preto\align{\small}
\expandafter\preto\csname align*\endcsname{\small}
\preto\equation{\par\nobreak\small\noindent}

\usepackage{enumitem}

\definecolor{LightCyan}{rgb}{0.92,1,1}
\title{Exploring a Unified Vision-Centric Contrastive Alternatives on Multi-Modal Web Documents}

\author{Yiqi Lin${^1}$ \quad Alex Jinpeng Wang$^{2}$ \quad Linjie Li$^{3}$ \quad Zhengyuan Yang$^{3}$  \quad \bf Mike Zheng Shou$^{1}$\\[3pt]
$^1$Show Lab, National University of Singapore \quad $^2$Central South University \quad $^3$Microsoft}
\begin{document}

\maketitle

\begin{abstract}
Contrastive vision-language models such as CLIP have demonstrated strong performance across a wide range of multimodal tasks by learning from aligned image-text pairs. However, their ability to handle complex, real-world web documents remains limited, particularly in scenarios where text and images are interleaved, loosely aligned, or embedded in visual form. To address these challenges, we propose Vision-Centric Contrastive Learning (\Mname), a unified framework that models text, images, and their combinations using a single vision transformer. \Mname~operates entirely in pixel space by rendering all inputs, whether textual, visual, or combined, as images, thus eliminating the need for OCR, text tokenization, or modality fusion strategy. To capture complex cross-modal relationships in multimodal web documents, \Mname~employs a snippet-level contrastive learning objective that aligns consecutive multimodal segments, leveraging the inherent coherence of documents without requiring explicitly paired image-text data.
To assess the effectiveness of this approach, we introduce three retrieval benchmarks, AnyCIR, SeqCIR, and CSR, designed to evaluate cross-modal retrieval, fine-grained sequential understanding, and generalization to unseen data, respectively. Empirical results show that \Mname~achieves competitive or superior performance compared to CLIP-style models on both the proposed benchmarks and established datasets such as M-BEIR and MTEB. These findings underscore the potential of multimodal web data as a valuable training resource for contrastive learning and illustrate the scalability of a unified, vision-centric approach for multimodal representation learning. Code and models are available at: \hyperlink{blue}{https://github.com/showlab/VC2L}.
\end{abstract}

\section{Introduction}

\begin{wrapfigure}{r}{0.5\textwidth}
  \begin{center}
\vspace{-20pt}
\includegraphics[width=0.5\textwidth]{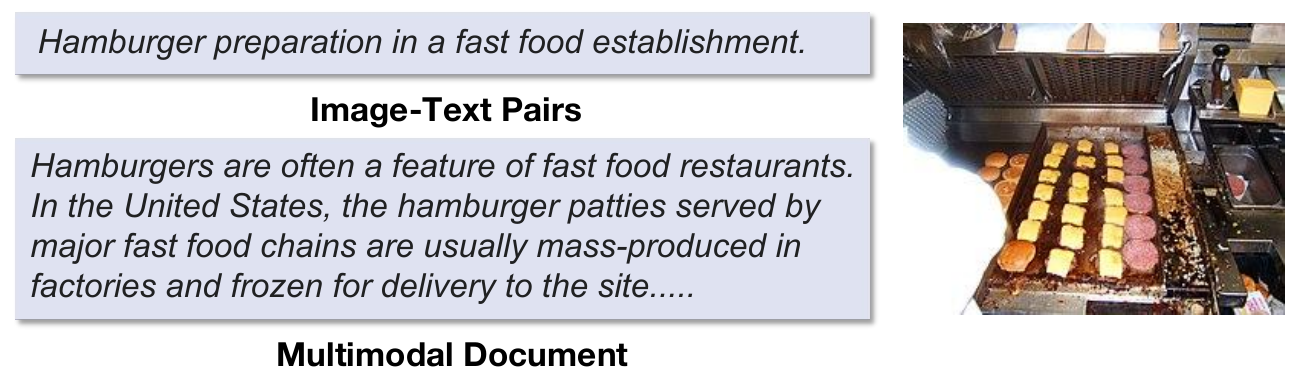}
  \end{center}
  \vspace{-10pt} 
  \caption{A comparison of image-caption pairs (Alt-Text) and multimodal documents (Wikipedia).}
  \label{fig:demo}
  \vspace{-5pt}
\end{wrapfigure}
Learning vision-language correspondence from image-text pairs has significantly advanced multi-modal research, with the rise of contrastive learning methods like CLIP~\citep{radford2021learning}. These models~\citep{radford2021learning,jia2021scaling,li2022blip,zhai2023sigmoid,sun2023eva} align vision and language representations within a shared space and demonstrate strong zero-shot capabilities across a range of downstream tasks~\citep{gu2021open,minderer2022simple, ramesh2021zero,ding2022decoupling,wortsman2022robust}. 

Despite their impressive performance, CLIP-style models face notable challenges when applied to real-world multimodal document~\cite{zhu2024multimodal,laurenccon2024obelics,awadalla2024mint} scenarios, e.g., retrieval, which often feature long-form content with interleaved text and images. Such scenarios reveal several key limitations in existing models. 
First, they struggle with interleaved multimodal inputs, where either the query or the retrieval target, or both, may contain combinations of text and images. Handling such inputs often requires additional post-processing or cross-modality fusion strategies~\citep{wei2023uniir}. Second, these models assume direct access to text, which is not always available in formats like scanned documents or image-based PDFs, where text is embedded as pixels and requires OCR for extraction. Finally, real-world documents are often long-form and loosely aligned across modalities, unlike datasets such as MS-COCO~\citep{chen2015microsoft} or LAION~\citep{schuhmann2022laion}, which provide clear correspondence between image-text pairs.
In practice, documents frequently contain semantically related but unpaired elements, for instance, a paragraph may be followed by a relevant image without explicit correspondence or linkage, as shown in Figure~\ref{fig:demo}. This setting differs significantly from the standard image-caption style data.

In this paper, we seek to explore the potential of  \textit{\textbf{directly training CLIP on multi-modal interleaved documents to overcome these challenges}}, given its foundational role in shaping vision-language learning.
To address these challenges, we propose Vision-Centric Contrastive Learning (\Mname), a unified framework that processes all input modalities (text, images, and interleaved content) directly in pixel space. Inspired by CLIPPO~\citep{tschannen2023clippo}, \Mname~renders both textual and visual information as images and processes them with a single vision transformer. Input content is organized into a $2\times2$ visual grid, which may contain image-only, text-only, or combined elements, as shown in Figure~\ref{fig:intro}. This unified vision-centric approach eliminates the need for separate encoders, text tokenization, or OCR, and seamlessly accommodates diverse modality input forms.

Beyond input space and model unification, \Mname~introduces a snippet-level contrastive learning strategy that leverages the natural coherence of document content.
Rather than depending on explicitly aligned image-text pairs, our approach samples consecutive multimodal snippets from the same document and encourages their embeddings to be similar.
Although these snippets are not strictly aligned, their sequential positioning often mirrors how humans interpret multimodal narratives, enabling a scalable and efficient solution for modeling interleaved real-world documents.
Furthermore, we propose modality masking and text masking augmentation to diversify the contrastive target by randomly masking portions of the content within sampled multimodal snippets.

To evaluate the capacity of \Mname~learn from multi-modal web documents, we design AnyCIR benchmark to evaluate the any-to-any modality information retrieval and SeqCIR benchmark to assess the fine-grained consecutive relationship modeling within documents by retrieving consecutive snippets sequentially.
To evaluate the transferability of \Mname~in real-world scenarios, we further design a zero-shot consecutive slide retrieval (CSR) benchmark, where slides are more complex image-text interleaved data.
Our extensive experiments also show that \Mname~can achieve superior zero-shot multi-modal information retrieval on M-BEIR~\citep{wei2023uniir} and text embedding learning on MTEB~\citep{muennighoff2023mteb}.
Additionally, we also investigate the impact of various contrast targets (image-caption, consecutive and non-consecutive snippets) and observe that joint image-text interleaved training can further improve language understanding in pixel space.

\textbf{Contributions:}
1). To the best of our knowledge, \Mname~is the first CLIP-style framework trained directly on image-text interleaved web documents, which opens new opportunities for leveraging large-scale, loosely aligned multimodal content as training data.
2). \Mname~is a single unified vision transformer operating in pixel space to handle text, images, and interleaved inputs, enabling effective multimodal understanding without OCR, tokenization, or modality-specific encoders.
3). To facilitate the evaluation of diverse modality understanding, we propose three consecutive information retrieval benchmarks, including AnyCIR, SeqCIR, and CSR.
Moreover, our extensive experimental results show that \Mname~achieves superior performance in our proposed benchmarks, the zero-shot multi-modal information retrieval benchmark M-BEIR, and the text embedding benchmark MTEB.

\section{Related Work}
\subsection{Vision-Language Learning from Web Data}
The pioneer work CLIP~\citep{radford2021learning} establishes a breakthrough learning paradigm by applying contrastive learning on large-scale noisy image/alt-text paired data from the internet.
Follow-up studies scale the image-text pairs data~\citep{schuhmann2022laion,gadre2024datacomp} and the model design~\citep{li2022blip,yu2022coca,zhai2023sigmoid} to further improve the performance.
More recently, with the rapid development of Multi-modal Large Language Models (MLLMs)~\citep{li2023blip,liu2024visual,lin2024vila}, multi-modal web documents data, such as MMC4~\citep{zhu2024multimodal} and OBELICS~\citep{laurenccon2024obelics}, have emerged as new sources of training data.
These multi-modal documents typically consist of sequences of coherent text paragraphs interleaved with images.
Several research~\citep{lin2024vila,mckinzie2024mm1} demonstrate that joint training with image-text data and multi-modal web documents outperforms solely image-text pairs, which indicates the multi-modal documents contain useful vision-language correspondence from image-text pairs.
Moreover, recent studies have explored advanced multi-modal embeddings across different text sources~\citep{wei2023uniir,jang2024mate}, improved long-form caption handling~\citep{zhang2024long,zheng2024dreamlip,zhang2024assessing} and leveraging MLLMs~\citep{ma2024unifying,lu2024text,jang2024mate,jiang2024e5,zhang2024gme,jiang2024vlm2vec,wang2024leveraging,zhou2024vista,lyu2025pixelworld,gu2025breaking} to encode multi-modal information for question answering or retrieval.
\textit{Differently, our goal is to offer a complementary perspective by exploring the potential of training a vision-centric CLIP model on multimodal web data, which presents new opportunities for a more versatile vision backbone in future MLLM pipelines.}

\subsection{Visual Representation for Language Modeling}
Despite the impressive results achieved by text tokenization~\citep{devlin2018bert,sennrich2015neural} in language modeling~\citep{devlin2018bert,brown2020language}, text tokenization is vulnerable to text permutations~\citep{salesky2021robust}, such as misspellings and has limited scalability to other languages~\citep{rust2022language}.
To address these challenges, a line of work explores the tokenizer-free solution based on the visual representation of text.
\citep{meng2019glyce} uses glyph-vectors from Chinese character images to enhance the text representation.
\citep{salesky2021robust} proposed visual text representation as open-vocabularies to improve the robustness of machine translation.
Recently, to close the gaps between the visual text representation and text tokenization, \citep{rust2022language,xiao2024pixel,gao2024improving,chai2024dual,wang2025textatlas5m} further explore different pre-training strategies on visual text images, such as next patch prediction, next token prediction, and contrastive learning.
In the vision-language domain, the most closely related work is CLIPPO~\citep{tschannen2023clippo}.
CLIPPO utilizes rendered alt-text and image pairs to train the vision encoder using contrastive learning, the same as CLIP. 
\textit{In contrast, \Mname marks the first attempt at exploration in the new source of training data, i.e., multimodal interleaved documents.}
Additionally, screenshot understanding~\citep{gao2024improving,liu2025any} is also closely related to visual text representation learning, which involves language modeling from documents~\citep{kim2022ocr}, web pages~\citep{lee2023pix2struct}, or UI images~\citep{li2022spotlight}.
Despite directly learning text information from images, these screenshot language models can not handle omni-modality input.

\section{Methodology}

\begin{figure*}
\centering
\includegraphics[width=\linewidth]{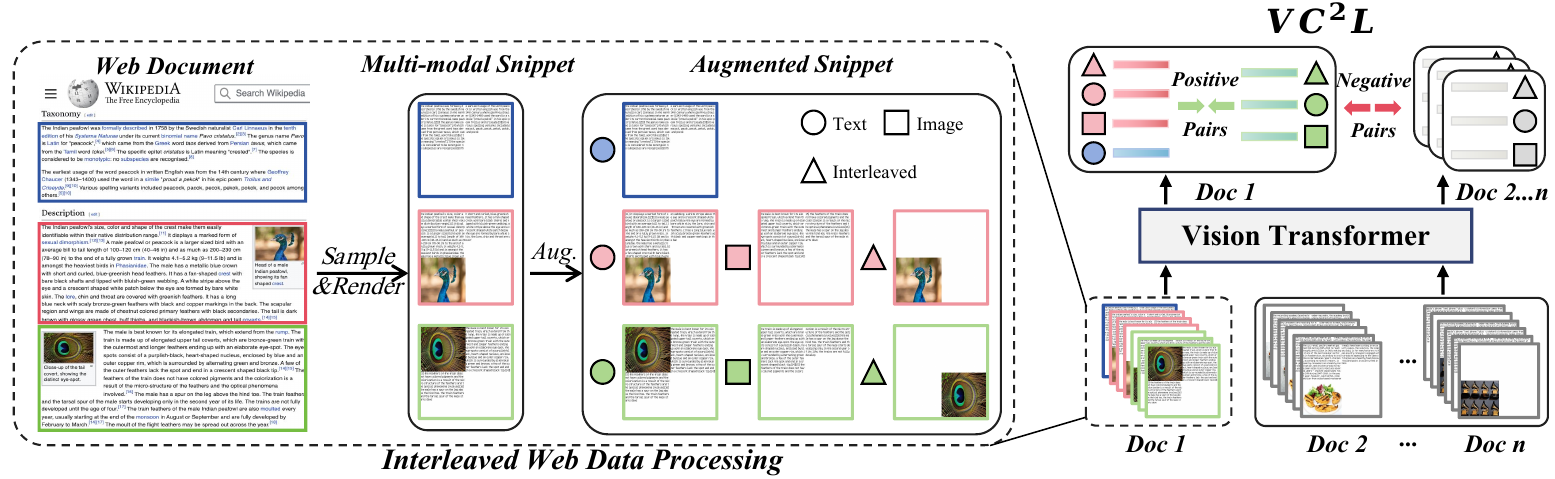}
\vspace{-15pt}
\caption{\Mname~explore an alternative vision-centric paradigm for unified vision-language modeling on interleaved web data. A \textit{single} vision transformer is used to process any image-text modality from pixels and thereby natively learn a unified representation.}
\label{fig:intro}
\vspace{-15pt}
\end{figure*}

As shown in Fig.~\ref{fig:intro},~\Mname~uses rendered consecutive snippets sampled from multi-modal web documents as training data.
After data pre-processing and augmentation, each snippet in positive pairs can be either image-only, text-only, or an interleaved image-text rendered image.
During training, the single vision model is optimized by contrastive loss on these consecutive data pairs.

\subsection{Interleaved Web Data Processing}
\label{sec:data_preprocess}
\textbf{Document Pre-processing.}
Given a web document, our goal is to sample a pair of semantically relevant image-text snippets for training.
Firstly, we split a document text into multiple text segments with a maximum of 1,100 characters in each segment.
Then, we leverage the image assignment annotation provided in MMC4 dataset~\citep{zhu2024multimodal} to assign the image to its corresponding segments.
Each interleaved snippet at least contains text but can be without images or assigned multiple images.
For the multiple image cases, we only randomly sample one image for training.

\noindent\textbf{Data Augmentation.}
Next, we apply two types of augmentations to obtain augmented snippets, i.e., \textit{modality masking} and \textit{text masking}.
In modality masking, we only mask snippets with both text and image contents.
During training, we apply modality masking with a masking rate of 40\% on snippets to randomly drop one modality content.
With modality masking, we are able to sample diverse training matching targets.
For text masking, we randomly remove sentences from the beginning or end of the text content in 40\% of the snippets. 
Note that text masking is only applied to snippets containing more than four sentences.
This augmentation enhances the model’s language understanding by preventing the model from overfitting recurring words.

\noindent\textbf{Multi-modal Snippet Rendering.}
Given a multimodal snippet containing both image and text, we render its content into a 2$\times$2 grid. Each grid has a resolution of 224$\times$224 pixels.
If the snippet includes an image, we resize it to fit the grid and place it in a randomly selected grid cell. 
For visual text rendering, we follow the approach in~\citep{tschannen2023clippo} using the GNU Unifont bitmap font.
The long-form text can be rendered across multiple grids, starting from the top-left and proceeding left-to-right and top-to-bottom. 
Once one grid is fulfilled with either image or text content, the rendering process continues in the next available grid.
More details are provided in the supplementary material. 

\vspace{-2pt}
\subsection{Training Objectives}

\textbf{Positive Pairs Sampling.}
After data pre-processing, a document $d_i$ is segmented as a serials of snippets, i.e., $\{s_{i}^{n}\}_{n=0}^N \in d_i$.
During training, we sample snippet pairs $(s^{q}_i, s^{k}_i)$ from the same documents $d_i$ as positive pairs, while the snippets from other documents are negative terms.
We use consecutive snippets, i.e., $k= q+1$, to construct positive pairs as our default setting.
To ablate the optimal training targets, we also investigate the sampling strategy of pairs with one-hop distance, i.e., $k = q+2$.
To differentiate, we use \textit{\textbf{Omni\footnote{In this paper, Omni denotes the image, text, and image-text interleaved modality }}} to denote consecutive pairs only, and \textit{\textbf{Omni+/++}} to denote 20\%/40\% of pairs are sampled from one-hop distance pairs. 

\noindent\textbf{Contrastive Learning.} Our training objective is contrastive loss~\citep{oord2018representation} formulated as,

\begin{equation} 
\label{eq:1}
 \mathcal{L}_c  =  -\frac{1}{N}\sum_{i=1}^{N}{log \frac{exp(f^{q}_{i}\cdot f^{k}_{i}) / \tau )}{\sum_{j=1}^{N} exp(f^{q}_{i}\cdot f^{k}_{j}) / \tau) }},
\end{equation}
where $(f^{q}_{i}, f^{k}_{i})$ is the visual features extracted from sampled snippets $(s^{q}_i, s^{k}_i)$ from the same document $d_i$ and temperature $\tau$ controls the sharpness of the logit distribution.

\vspace{-5pt}
\section{Consecutive Information Retrieval}

To evaluate the consecutive information retrieval capabilities, we design two multi-modal snippet retrieval benchmarks based on OBELICS~\citep{laurenccon2024obelics} and zero-shot slide retrieval based on Slideshare-1M~\citep{araujo2016large}.
Compared to the training dataset MMC4, the OBELICS preserves the original image text interleaved order, which is closer to real-world scenes.
The slides in Slidershare-1M are naively interleaved multi-modal data with more complex interleaved forms.

\begin{figure}
    \centering
    \includegraphics[width=\linewidth]{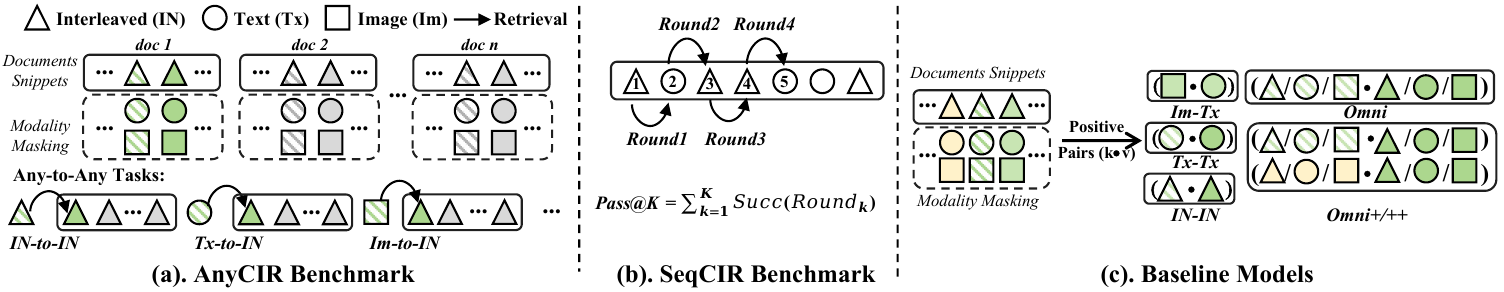}
    \vspace{-12pt}
    \caption{\textbf{(a):} In AnyCIR, we first sample consecutive snippet pairs from distinct documents and use the former snippet to retrieve the latter one. For each query, all the later snippets are candidates.
    \textbf{(b):} In SeqCIR, we sequentially retrieve the consecutive snippets in multiple rounds.
    For each query, all the snippets segmented from sampled documents are candidates while ignoring preceding snippets from the previous round.
    \textbf{(c):} The positive contrastive pair settings of different baseline models.
    }
    \label{fig:setting}
\vspace{-5pt}
\end{figure}

\noindent\textbf{Any-to-Any Consecutive Information Retrieval (AnyCIR).} 
In this task, we aim to retrieve any modality consecutive information given any modality queries, as shown in Fig.~\ref{fig:setting}a.
The types of modality include interleaved (\textbf{IN}), Text only (\textbf{Tx}), and Image only (\textbf{Im}), resulting in 9 tasks in total with different combinations.
The AnyCIR consists of 20,000 randomly sampled consecutive snippet pairs from distinct documents.
Each snippet in the pair includes text and at least one image content.
During inference, all the tasks share the same snippet pair source.
For retrieval tasks with a single modality, we simply mask other modalities during rendering.
We render images into a randomly chosen grid for both queries and candidates.

\noindent\textbf{Sequential Consecutive Information Retrieval (SeqCIR)}.
This task aims to evaluate the fine-grained consecutive information modeling capacity.
For each query, the candidate pool consists of 26,433 snippets from 5,000 distinct documents.
For each snippet, we use the full text and one randomly selected image if applicable.
We use 2,524 snippets as the initial query set, which are the first snippets of the documents.
For this task, we iteratively retrieve the next consecutive snippets and only successful retrieval queries are passed to the next iteration.
For each iteration, we ignore the preceding snippets ahead of the query snippet in the documents.
The Pass@K rate denotes the success rate of sequential retrieval at the $k^{th}$ round, as shown in Fig.~\ref{fig:setting}b.
The SeqCIR is a very challenging task as the candidate pool of SeqCIR contains subsequent snippets from the same documents. It requires the model to accurately distinguish the most consecutive snippet.

\noindent\textbf{Zero-Shot Consecutive Slide Retrieval (CSR).}
To better examine the transferability of~\Mname~under real-world scenario, we propose a benchmark of retrieving the most relevant slide.
Specifically, we sample 28,016 pairs of consecutive slide images from Slideshare-1M~\citep{araujo2016large}.
Each pair is sampled from a distinct slide deck (more than 6 slides) after removing the first two slides.
For evaluation, we use the former slide as a query and all the latter slides as candidates.
Despite consecutive slides might share similar layouts or part of the content overlap, our experimental results show that it is still a challenging task even using these shortcuts instead of understanding the multimodal information.

\section{Experiments}
\subsection{Experimental Setup.}

\textbf{Data Variant Baselines.}
To better understand the model capacity learned from interleaved data, we further construct different positive pair data as our baselines as illustrated in Fig.~\ref{fig:setting}c. Our baselines include \textbf{1).} Image-Text \textbf{(Im-Tx)} pairs sampled from a LAION subset;
\textbf{2).} Image-Text \textbf{(Im-Tx)} pairs from the same snippet of MMC4, where we use the MMC4 annotation to generate the pairs, i.e. the CLIP similarity assignment;
\textbf{3).} Text-Text \textbf{(Tx-Tx)} pairs by masking all the images in the snippets;
\textbf{4).} Interleaved-Interleaved \textbf{(IN-IN)} pairs by sampling from the snippets pairs containing both image and text content;
\textbf{5).} \textbf{Omni$_{224}$} pairs first rendering in 448 $\times$ 448 resolution then resize to 224  $\times$ 224 resolution for fair comparison with original CLIP model;
\textbf{6).} \textbf{Omni+/++} denotes 20\%/40\% of pairs are sampled from one-hop pairs.
All baselines use the same training setting.

\noindent\textbf{Implementation Details.}
Our implementation is based on OpenCLIP~\citep{openclip}.
In all experiments, we use ViT-B-16~\citep{dosovitskiy2020image} with an input resolution 448$\times$448.
We use a batch size of 1024 and a learning rate of 1e-4 for training 20 epochs.
Our pretraining dataset uses the MMC4-core-fewer-face~\citep{zhu2024multimodal} subset, comprising 5 million documents with both images and text, totaling 17 million images.
We use CLIP~\citep{radford2021learning} checkpoint as our initialization due to the small scale of our training data. 
We include the vision encoder of CLIP~\citep{radford2021learning}, OpenCLIP~\citep{cherti2023reproducible}, and CLIPPO~\citep{tschannen2023clippo} in the model size of ViT-B as our baseline.
Note that these baselines are trained on different sources and scales of image-text pair data.

\subsection{Consecutive Multi-Modal Retrieval}

\textbf{Any-to-Any Consecutive Information Retrieval (AnyCIR).}
In Table~\ref{tab:ocir}, we report 9 retrieval task results at Rank@1 metric.
It can be observed that image-text interleaved data can help the model better understand visual text data.
For example, Omni and IN-IN models achieve better results on the Tx-to-Tx retrieval task than the Tx-Tx baseline.
Moreover, more diverse training data can boost the performance of omni-modality representation learning, as Omni achieves better performance on the IN-to-IN task compared to the IN-IN baseline.
When training the model with non-consecutive samples, i.e., Omni+ or Omni++, the performance only slightly decreases, which indicates that the close snippets generally have consistent vision-language correspondence.
Additionally, Omin$_{224}$ indicates that our performance gains are not only from the higher input resolution but also from our novel training data design.
Interestingly, the CLIP vision encoder has stronger visual text understanding capacity over OpenCLIP, which is trained on a larger scale of datasets.
When training on image-text pair data from LAION, the model performs poorly on the AnyCIR benchmark, indicating the large domain gap between image-caption and multi-modal document data.

\begin{table*}
\caption{Any-to-Any Consecutive Information Retrieval benchmark on Rank@1 metric. The modalities include Image-Text Interleaved (\textbf{IN}), Text only (\textbf{Tx}), and Image only (\textbf{Im}). \colorbox[HTML]{EFEFEF}{Gray} results refer to the model input resolution as 224 and the default is 448.
}
\centering
\tablestyle{4pt}{1} 
\resizebox{\linewidth}{!}{%
\begin{tabular}{l|c|ccccccccc|c}
\hline
Model & Data & IN-IN & IN-Tx & IN-Im & Tx-IN & Tx-Tx & Tx-Im& Im-IN & Im-Tx & Im-Im & Overall \\ \hline
\rowcolor[HTML]{EFEFEF} 
CLIP-V\citep{radford2021learning} & WIT 400M\citep{radford2021learning} & 24.10 & 6.18 & 5.27 & 14.23 & 11.47 & 1.02 & 11.60 & 0.93 & 12.45 & 9.69 \\
\rowcolor[HTML]{EFEFEF} 
OpenCLIP-V\citep{openclip} & LAION 2B\citep{schuhmann2022laion} & 18.41 & 0.26 & 12.23 & 4.73 & 3.82 & 0.86 & 13.52 & 0.02 & 15.76 & 7.73 \\
\rowcolor[HTML]{EFEFEF} 
CLIPPO\citep{tschannen2023clippo} & YFCC 100M\citep{thomee2016yfcc100m} & 10.17 & 0.01 & 9.99 & 0.00 & 0.01 & 0.01 & 6.31 & 0.02 & 11.79 & 4.25 \\
\rowcolor[HTML]{EFEFEF} 
\Mname~(Omni$_{224}$) & MMC4-core\citep{zhu2024multimodal} &
69.39& 67.20& 13.89& 67.86& 70.61& 5.04& 14.00& 5.68& 14.45& 36.45 \\ \hline
\Mname~(Im-Tx) & LAION 40M\citep{schuhmann2022laion} & 25.64 & 15.23 & 11.89 & 21.21 & 26.40 & 5.72 & 15.07 & 5.36 & 16.20 & 15.86 \\
\Mname~(Im-Tx) & MMC4-core\citep{zhu2024multimodal} & 63.34 & 59.15 & 15.60 & 61.30 & 61.08 & \textbf{12.34} & 17.36 & \textbf{12.31} & 17.97 & 35.60 \\
\Mname~(Tx-Tx) & MMC4-core\citep{zhu2024multimodal} & 53.16 & 62.34 & 0.01 & 61.12 & 73.38 & 0.01 & 0.03 & 0.02 & 0.78 & 27.87 \\
\Mname~(IN-IN) & MMC4-core\citep{zhu2024multimodal} & 76.56 & \textbf{74.85} & 0.40 & \textbf{74.19} & \textbf{74.81} & 0.12 & 2.58 & 0.64 & 8.95 & 34.79 \\ \hline
\Mname~(Omni) & MMC4-core\citep{zhu2024multimodal} & \textbf{78.27} & 73.89 & \textbf{22.10} & \textbf{74.19} & 74.32 & 10.08 & \textbf{22.00} & 10.95 & {19.50} & \textbf{42.81} \\
\Mname~(Omni+) & MMC4-core\citep{zhu2024multimodal} & 77.94 & 73.68 & 21.87 & 73.73 & 73.68 & 10.06 & 21.76 & 10.70 & 19.29 & 42.52 \\
\Mname~(Omni++) & MMC4-core\citep{zhu2024multimodal} & 78.05 & 73.53 & 21.27 & 73.57 & 73.41 & 9.96 & 21.48 & 10.63 & \textbf{19.55} & 42.38 \\ \hline
\end{tabular}
}
\vspace{-5pt}
\label{tab:ocir}
\end{table*}

\begin{table}
\caption{Sequential Consecutive Information Retrieval. Pass@k denotes the retrieval success rate at $k^{th}$ round. \colorbox[HTML]{EFEFEF}{Gray} results refer to the model input resolution as 224 and the default is 448.}
\centering
\tablestyle{5pt}{1} 
\resizebox{0.7\linewidth}{!}{
\begin{tabular}{l|c|cccc}
\hline
Model & Data & Pass@1 & Pass@2 & Pass@3 & Pass@4 \\ \hline
\rowcolor[HTML]{EFEFEF} 
CLIP-V\citep{radford2021learning} & WIT 400M\citep{radford2021learning} & 11.69 & 1.51 & 0.24 & 0.04 \\
\rowcolor[HTML]{EFEFEF} 
OpenCLIP-V\citep{openclip} & LAION 2B\citep{schuhmann2022laion} & 7.49 & 0.71 & 0.16 & 0.00 \\
\rowcolor[HTML]{EFEFEF} 
CLIPPO\citep{tschannen2023clippo} & YFCC 100M\citep{thomee2016yfcc100m} & 3.86 & 0.36 & 0.09 & 0.00 \\
\rowcolor[HTML]{EFEFEF} 
\Mname~(Omni$_{224}$) & MMC4-core\citep{zhu2024multimodal} & 31.85 & 10.97 & 5.39 & 2.81 \\ \hline
\Mname~(Im-Tx) & LAION 40M\citep{schuhmann2022laion} & 13.00 & 1.90 & 0.32 & 0.04 \\
\Mname~(Im-Tx) & MMC4-core\citep{zhu2024multimodal} & 29.48 & 9.03 & 3.80 & 1.58 \\
\Mname~(Tx-Tx) & MMC4-core\citep{zhu2024multimodal} & 26.39 & 7.21 & 3.01 & 1.55 \\
\Mname~(IN-IN) & MMC4-core\citep{zhu2024multimodal} & 32.53 & {12.96} & 6.38 & 3.57 \\ \hline
\Mname~(Omni) & MMC4-core\citep{zhu2024multimodal} & \textbf{34.43} & \textbf{13.07} & \textbf{6.78} & \textbf{3.76} \\
\Mname~(Omni+) & MMC4-core\citep{zhu2024multimodal} & 33.28 & 12.60 & 6.50 & 3.68 \\
\Mname~(Omni++) & MMC4-core\citep{zhu2024multimodal} & 33.76 & 12.56 & 6.42 & 3.76 \\ \hline
\end{tabular}
}
\vspace{-10pt}
\label{tab:scir}
\end{table}

\begin{table}
\caption{Zero-Shot Consecutive Slides Retrieval. \colorbox[HTML]{EFEFEF}{Gray} results refer to the model input resolution as 224 and the default is 448.
}
\tablestyle{6pt}{1} 
\centering
\resizebox{0.65\linewidth}{!}{%
\begin{tabular}{l|c|ccc|c}
\hline
Model & Data & R@1 & R@5 & R@10 & Avg \\ \hline
\rowcolor[HTML]{EFEFEF} 
CLIP-V\citep{radford2021learning} & WIT 400M\citep{radford2021learning} & 34.60 & 45.10 & 49.29 & 43.00 \\
\rowcolor[HTML]{EFEFEF} 
OpenCLIP-V\citep{openclip} & LAION 2B\citep{schuhmann2022laion} & \textbf{38.08} & \textbf{48.33} & \textbf{52.27} & \textbf{46.23} \\
\rowcolor[HTML]{EFEFEF} 
CLIPPO\citep{tschannen2023clippo} & YFCC 100M\citep{thomee2016yfcc100m} & 26.42 & 34.31 & 37.30 & 32.68 \\
\rowcolor[HTML]{EFEFEF} 
\Mname~(Omni$_{224}$) & MMC4-core\citep{zhu2024multimodal} & 33.81 & 43.28 & 47.02 & 41.37 \\ \hline
\Mname~(Im-Tx) & LAION 40M\citep{schuhmann2022laion} & 26.21 & 33.13 & 35.85 & 31.73 \\
\Mname~(Im-Tx) & MMC4-core\citep{zhu2024multimodal} & 34.68 & 43.45 & 46.85 & 41.66 \\
\Mname~(Tx-Tx) & MMC4-core\citep{zhu2024multimodal} & 11.04 & 14.59 & 16.14 & 13.92 \\
\Mname~(IN-IN) & MMC4-core\citep{zhu2024multimodal} & 25.92 & 33.40 & 36.46 & 31.93 \\ \hline
\Mname~(Omni) & MMC4-core\citep{zhu2024multimodal} & 44.05 & \textbf{55.55} & \textbf{59.74} & 53.11 \\
\Mname~(Omni+) & MMC4-core\citep{zhu2024multimodal} & \textbf{44.21} & 55.54 & 59.68 & \textbf{53.14} \\
\Mname~(Omni++) & MMC4-core\citep{zhu2024multimodal} & 43.74 & 55.16 & 59.29 & 52.73 \\ \hline
\end{tabular}
}

\label{tab:ppt}
\end{table}

\noindent\textbf{Sequential Consecutive Information Retrieval (SeqCIR)}.
Table~\ref{tab:scir} reports sequential consecutive snippets retrieval results in a total of four rounds.
The best model only achieves a 3.7\% success rate after four rounds, which indicates that these models still lack of capacity for fine-grained consecutive relation modeling.
The results also draw the same observation as the AnyCIR benchmark that diverse training data helps omni-modality representation learning.

\noindent\textbf{Zero-Shot Consecutive Slide Retrieval (CSR).}
As shown in Table~\ref{tab:ppt}, the Omni model achieves the best results with 44\% rank@1 accuracy under zero-shot setting.
It indicates that our learned interleaved representation is able to generalize to the complex interleaved data, i.e. slide.
Moreover, the results demonstrate that the language understanding capacity of ~\Mname can be generalized beyond rendered text to various styles and font sizes.
We also find that OpenCLIP is better than CLIP in CSR, which contrasts with previous benchmarks.
One possible reason is that the OpenCLIP has been trained with slide data as shown in ~\citep{lin2023parrot}.

\begin{table*}[]
\caption{Zero-shot results on M-BEIR$_{union}$ (Recall@5). Im-Tx$_{la}$ denotes training on LAION data.}
\tablestyle{1pt}{1} 
\resizebox{\linewidth}{!}{%
\begin{tabular}{ll|ccccc|cccc|ccc}
\hline
{Task} & {Dataset} & CLIP$_{B}$\citep{radford2021learning} & CLIP$_{L}$\citep{radford2021learning} & SigLIP\citep{zhai2023sigmoid} & BLIP\citep{li2022blip} & BLIP2\citep{li2023blip} & Im-Tx$_{la}$ & Im-Tx & Tx-Tx & IN-IN & Omni & Omni+ & Omni++ \\ \hline
\multirow{3}{*}{1. $q_t \to c_i$} & VisualNews & 0.0 & 0.0 & 0.0 & 0.0 & 0.0 & 0.2 & 0.1 & 0.0 & 0.0 & 0.2 & 0.2 & 0.2 \\
 & MSCOCO & 0.0 & 0.0 & 0.0 & 0.0 & 0.0 & 0.1 & 0.0 & 0.0 & 0.0 & 0.0 & 0.0 & 0.0 \\
 & Fashion200K & 0.0 & 0.0 & 0.0 & 0.0 & 0.0 & 0.1 & 0.0 & 0.0 & 0.0 & 0.0 & 0.0 & 0.0 \\ \hline
2. $q_t \to c_t$ & WebQA & 32.5 & 32.1 & 34.0 & 38.1 & 35.2 & 35.9 & 47.3 & 41.0 & 46.0 & 46.2 & 48.5 & 49.3 \\ \hline
\multirow{2}{*}{\shortstack[l]{3. $q_t$ \\ $\to$ ($c_i, c_t$)}} & EDIS & 3.0 & 6.7 & 1.1 & 0.0 & 0.0 & 1.7 & 2.3 & 4.4 & 11.4 & 10.6 & 11.5 & 12.3 \\
 & WebQA & 0.8 & 5.5 & 2.1 & 0.0 & 0.0 & 1.2 & 6.8 & 24.0 & 40.7 & 27.4 & 29.1 & 29.5 \\ \hline
\multirow{3}{*}{4. $q_i \to c_t$} & VisualNews & 0.0 & 0.0 & 0.0 & 0.0 & 0.0 & 0.0 & 0.2 & 0.0 & 0.0 & 0.2 & 0.3 & 0.2 \\
 & MSCOCO & 0.0 & 0.0 & 0.0 & 0.0 & 0.0 & 0.1 & 0.2 & 0.0 & 0.0 & 0.3 & 0.3 & 0.3 \\
 & Fashion200K & 0.0 & 0.0 & 0.0 & 0.0 & 0.0 & 0.0 & 0.0 & 0.0 & 0.0 & 0.0 & 0.0 & 0.0 \\ \hline
5. $q_i \to c_t$ & NIGHTS & 27.1 & 25.3 & 28.7 & 25.1 & 24.0 & 28.0 & 27.1 & 0.2 & 15.7 & 25.0 & 24.3 & 25.5 \\ \hline
\multirow{2}{*}{\shortstack[l]{6. ($q_i, q_t$) \\ $\to c_t$}} & OVEN & 0.0 & 0.0 & 0.0 & 0.0 & 0.0 & 0.0 & 0.3 & 0.0 & 0.1 & 0.6 & 0.6 & 1.0 \\
 & InfoSeek & 0.0 & 0.0 & 0.0 & 0.0 & 0.0 & 0.0 & 0.3 & 0.0 & 0.0 & 0.2 & 0.2 & 0.4 \\ \hline
\multirow{2}{*}{\shortstack[l]{7. ($q_i, q_t$) \\$\to c_i$}} & FashionIQ & 1.0 & 4.4 & 4.8 & 2.2 & 3.9 & 6.8 & 2.7 & 0.0 & 0.5 & 3.8 & 4.2 & 3.5 \\
 & CIRR & 1.6 & 5.4 & 7.1 & 7.4 & 6.2 & 7.4 & 3.1 & 0.0 & 0.2 & 5.5 & 5.9 & 5.7 \\ \hline
\multirow{2}{*}{\shortstack[l]{8. ($q_i, q_t$) \\$\to$ ($c_i, c_t$)}} & OVEN & 1.0 & 24.5 & 27.2 & 10.1 & 13.8 & 14.5 & 2.2 & 0.0 & 0.1 & 5.8 & 6.1 & 4.8 \\
 & InfoSeek & 0.6 & 22.1 & 24.3 & 7.9 & 11.4 & 11.1 & 1.7 & 0.0 & 0.2 & 4.2 & 4.6 & 3.1 \\ \hline
- & Average & 4.2 & 7.9 & 8.1 & 5.7 & 5.9 & 6.7 & 5.9 & 4.3 & 7.2 & 8.1 & \textbf{8.5} & \textbf{8.5} \\ \hline
\end{tabular}
}
\label{tab:m_beir}
\end{table*}

\begin{table*}[]
\caption{Mass Text Embedding Benchmark. The rows in \colorbox{LightCyan}{Cyan} refer to the text encoder directly processing the text input. \colorbox[HTML]{EFEFEF}{Gray} results refer to input resolution as 224, and the default is 448.}
\centering
\tablestyle{8pt}{1} 
\resizebox{\linewidth}{!}{%
\begin{tabular}{l|ccccccc|c}
\hline
 & Class. & Clust. & PairClass. & Rerank. & Retr. & STS & Summ. & Avg. \\ \hline
Num. Datasets & 12 & 11 & 3 & 4 & 15 & 10 & 1 & 56 \\ \hline
\rowcolor{LightCyan} 
Glove\citep{pennington2014glove} & 57.29 & 27.73 & 70.92 & 43.29 & 21.62 & 61.85 & 28.87 & 41.97 \\
\rowcolor{LightCyan} 
Komninos\citep{komninos2016dependency} & 57.65 & 26.57 & 72.94 & 44.75 & 21.22 & 62.47 & 30.49 & 42.06 \\
\rowcolor{LightCyan} 
BERT\citep{devlin2018bert} & 61.66 & 30.12 & 56.33 & 43.44 & 10.59 & 54.36 & 29.82 & 38.33 \\
\rowcolor{LightCyan} 
SimCSE-BERT-unsup\citep{gao2021simcse} & 62.5 & 29.04 & 70.33 & 46.47 & 20.29 & 74.33 & 31.15 & 45.45 \\ \hline
\rowcolor{LightCyan} 
CLIP-T\citep{radford2021learning} & 60.17 & 32.7 & 75.4 & 46 & 14.76 & 65.7 & 30.29 & 42.9 \\
\rowcolor{LightCyan} 
OpenCLIP-T\citep{openclip} & 59.2 & 36.61 & 72.43 & 47.91 & 28.05 & 70.43 & 26.57 & 47.76 \\ \hline
\rowcolor[HTML]{EFEFEF} 
CLIP-V\citep{radford2021learning} & 55.76 & 31.64 & 63.85 & 45.12 & 14.51 & 62.55 & 26.81 & 40.34 \\
\rowcolor[HTML]{EFEFEF} 
OpenCLIP-V\citep{openclip} & 49.4 & 23.85 & 56.55 & 42.05 & 11.75 & 54.6 & 28.57 & 34.71 \\ \hline
\Mname~(Im-Tx/LAION) & 49.04 & 27.67 & 67.34 & 43.67 & 16.49 & 65.26 & 29.74 & 39.27 \\
\Mname~(Im-Tx) & 52.46 & 34.48 & 70.67 & 47.19 & 19.58 & 65.27 & 30.64 & 42.62 \\
\Mname~(Tx-Tx) & 51.12 & 33.26 & 70.62 & 46.56 & 17.89 & 65.51 & 26.72 & 41.56 \\
\Mname~(IN-IN) & 53.83 & 35.13 & 73.27 & 48.03 & 20.59 & 68.48 & 29.31 & 44.06 \\ \hline
\Mname~(Omni) & 53.69 & 36.75 & 72.34 & 48.10 & 21.93 & 67.18 & 28.44 & 44.41 \\
\Mname~(Omni+) & 53.25 & 36.95 & 72.50 & 48.34 & 23.07 & 67.62 & 27.91 & \textbf{44.76} \\
\Mname~(Omni++) & 52.95 & 36.99 & 71.99 & 48.29 & 22.27 & 67.58 & 27.79 & 44.45 \\ \hline
\end{tabular}
}
\end{table*}

\subsection{Traditional Multi-modal Information Retrieval}
To investigate the ability of \Mname~in traditional information retrieval tasks, we adopt zero-shot M-BEIR~\citep{wei2023uniir} for evaluation, which assembles 10 diverse datasets from multiple domains with 8 distinct multi-modal retrieval tasks.
In our setting, we render all modality information (image and text) into a single image for all the queries and candidates without using instructions.
As we find out the balance of the modality information is critical to this task, we pad all the text input to 800 chars by repeating them.
We provide the ablation study results on supply materials.

Table~\ref{tab:m_beir} shows the zero-shot union candidate pool results of \Mname~and baselines, including CLIP$_B$(ViT-B), CLIP$_L$(ViT-L), SigLIP~\citep{zhai2023sigmoid}, BLIP~\citep{li2022blip} and BLIP2~\citep{li2023blip}.
\Mname~using single vision encoder outperforms the models with separate text encoder under the zero-shot setting, e.g.SigLIP.
Also, it can be seen that the models trained on interleaved data generally are good at WebQA~\citep{chang2022webqa} while performing poorly on InfoSeek~\citep{chen2023can} compared to the CLIP-style model.
It indicates that the interleaved data and image-caption data empower the model with different capacities.

\subsection{Text Embedding Benchmark}
To evaluate the language understanding capability, we use MTEB~\citep{muennighoff2023mteb} English subset, which comprises 7 different tasks in a total of 56 datasets.
During inference, we render all text into images and use the pooled representation as the text embedding.
We can observe that \Mname~achieve competitive performance against most of unsupervised baselines, including Glove~\citep{pennington2014glove}, Komninos~\citep{komninos2016dependency}, BERT~\citep{devlin2018bert} and SimCSE~\citep{gao2021simcse}, which are trained on a large language corpus.
When training with one-hop pair samples as the alignment target, our model achieves better performance.
Similar to the aforementioned findings, the MTEB benchmark shows that the multi-modal data helps the model to better learn language representation from pixels.
We also provide the results of the text(-T) and vision(-V) encoder performance of CLIP and OpenCLIP, where the vision encoder input is rendered text at 224 resolution size.
Interestingly, the text encoder of OpenCLIP outperforms all the unsupervised baselines while its vision encoder poorly understands the visual text information.

\subsection{Discussion: Benefits of Unified Pixels Space}
\label{sec:discuss}
\Mname~provide a more general-purpose vision-centric encoder that can seamlessly understand the image, visual text, and their relationship.
Unifying everything into pixels can reduce specialized design in separate encoder counterparts (e.g. CLIP), resulting in a much lower computational cost compared to forwarding text inputs through an additional text encoder or extracting text through OCR models.
Moreover, our approach supports a maximum text input length of 1,100 characters ($\approx$ 275 tokens) in a fixed cost, while the text input of CLIP is limited to 77 tokens.

\begin{figure*}
\includegraphics[width=\linewidth]{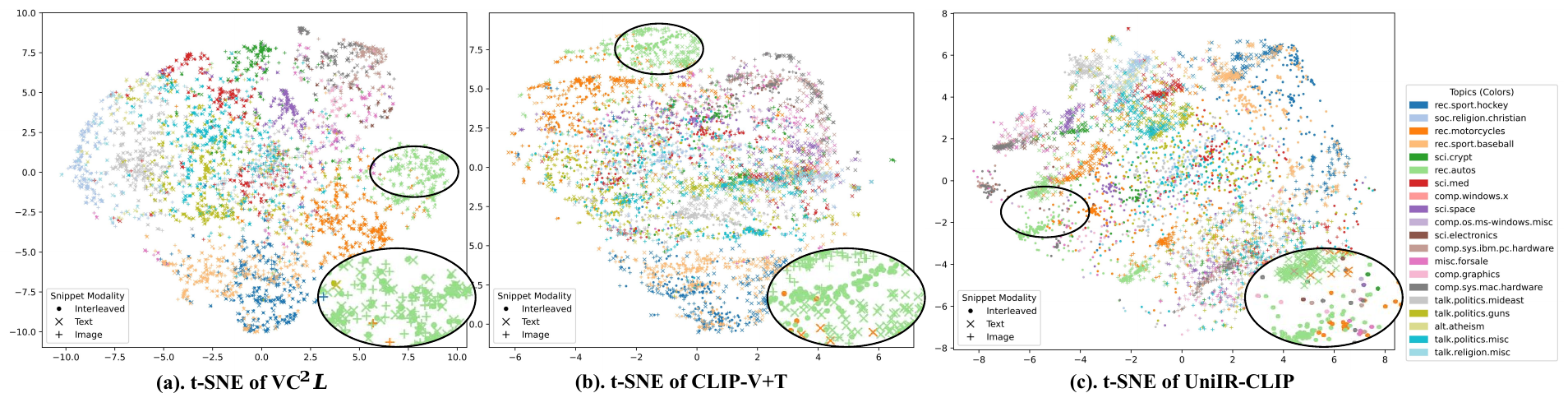}
\vspace{-15pt}
\caption{t-SNE visualization of interleaved, text and image snippets embedding on OBELICS.
}
\label{fig:tsne}
\end{figure*}

\noindent\textbf{Embedding Space.}
In Fig.~\ref{fig:tsne}, we visualize the distribution of interleaved, image and text embeddings from the same snippets of three models, including~\Mname, CLIP-V+T with averaging features, and UniIR-CLIP~\citep{wei2023uniir}.
The labels of the snippet are predicted by topic model~\citep{grootendorst2022bertopic} trained on 20NewsGroups~\citep{lang1995newsweeder}.
It can be observed that our model can learn useful representations that are aligned with linguistic semantics, as snippets on similar topics are close to each other.
Compared to the separate encoder baselines,~\Mname~learn a more unified omni-modality representation, which indicates that unifying in pixel space can further reduce the modality discrepancy.

\begin{table*}[]
\caption{Ablation experiments on AnyCIR benchmark. The Avg denotes 9 tasks average performance.}
\vspace{-5pt}
\centering
\begin{subtable}[c]{0.45\linewidth}
\caption{Model initialization.}
\vspace{-5pt}
\centering
\tablestyle{8pt}{1} 
\resizebox{\linewidth}{!}{%
\begin{tabular}{c|c|ccc|c}
\hline
Init & Model & IN-IN & Tx-Tx & Im-Im & Avg \\ \hline
 & IN-IN & 65.85 & 64.55 & 6.46 & 29.60 \\
$\checkmark$ & IN-IN & \textbf{76.56} & \textbf{74.81} & \textbf{8.95} & \textbf{34.79} \\ \hline
 & Omni & 62.30 & 61.22 & 12.18 & 30.42 \\
$\checkmark$ & Omni & \textbf{78.27} & \textbf{74.32} & \textbf{19.50} & \textbf{42.81} \\ \hline
\end{tabular}
}
\label{ab:model_init}
\end{subtable}
\hfill
\begin{subtable}[c]{0.4\linewidth}
\caption{Image Rendering Positions.}
\vspace{-5pt}
\centering
\tablestyle{8pt}{1} 
\resizebox{\linewidth}{!}{%
\begin{tabular}{c|ccc|c}
\hline
Position & Im-IN & Im-Tx & Im-Im & Avg \\ \hline
grid-0 & 22.07& 10.88& \textbf{19.53} & 42.84\\
grid-1 & \textbf{22.18}& \textbf{11.03}& 19.50 & \textbf{42.88}\\
grid-2 & 22.01& 10.91& 19.51 & 42.84\\
grid-3 & \textbf{22.18}& \textbf{11.03}& 19.43 & 42.82\\ \hline
\end{tabular}
}
\label{tab:ab_posgrid}
\end{subtable}
\begin{subtable}[c]{0.28\linewidth}
\caption{Modality Masking.}
\vspace{-5pt}
\tablestyle{4pt}{1} 
\centering
\resizebox{\linewidth}{!}{%
\begin{tabular}{c|ccc|c}
\hline
Ratio & IN-IN & Tx-Tx & Im-Im & Avg \\ \hline
0.0 & 76.56 & \textbf{74.81} & 8.95 & 34.79 \\
0.2 & 76.22 & 71.63 & \textbf{19.50} & 41.74 \\
0.4 & 77.41 & 72.39 & 19.30 & \textbf{41.98} \\
0.6 & 77.60 & 73.29 & 18.74 & 41.75 \\
0.8 & \textbf{78.00} & 73.96 & 17.06 & 40.80 \\
1.0 & 76.56 & {74.26} & 8.71 & 34.70 \\ \hline
\end{tabular}
}
\label{tab:ab_mm}
\end{subtable}
\hfill
\begin{subtable}[c]{0.28\linewidth}
\caption{Text Masking.}
\vspace{-5pt}
\centering
\tablestyle{4pt}{1} 
\resizebox{\linewidth}{!}{%
\begin{tabular}{c|ccc|c}
\hline
Ratio & IN-IN & Tx-Tx & Im-Im & Avg \\ \hline
0.0 & 77.41 & 72.39 & 19.30 & 41.98 \\
0.2 & \textbf{78.34} & 74.26 & 19.27 & 42.71 \\
0.4 & 78.27 & \textbf{74.32} & {19.50} & \textbf{42.81} \\
0.6 & 77.70 & 73.56 & 19.48 & 42.48 \\
0.8 & 77.85 & 73.32 & \textbf{19.58} & 42.42 \\
1.0 & 77.41 & 72.60 & 19.08 & 41.96 \\ \hline
\end{tabular}
}
\label{tab:ab_tm}
\end{subtable}
\hfill
\begin{subtable}[c]{0.32\linewidth}
\caption{Non-Consecutive Pair Sampling.}
\vspace{-5pt}
\centering
\tablestyle{4pt}{1} 
\resizebox{\linewidth}{!}{%
\begin{tabular}{c|ccc|c}
\hline
Ratio & IN-IN & IN-Tx & IN-Im & Avg \\ \hline
0 & \textbf{78.27} & \textbf{74.32} & 19.50 &  {\textbf{42.81}} \\
0.1 & 78.04 & 73.53 & \textbf{19.74} & 42.54 \\
0.2 (+) & 77.94 & 73.68 & 19.29 & 42.52 \\
0.3 & 78.13 & 73.65 & 19.31 & 42.44 \\
0.4 (++) & 78.05 & 73.41 & 19.55 & 42.38 \\
0.5 & 77.95 & 73.54 & 19.29 & 42.31 \\ \hline
\end{tabular}
}
\label{tab:ab_pospair}
\end{subtable}
\end{table*}

\begin{figure*}
\centering
\includegraphics[width=\linewidth]{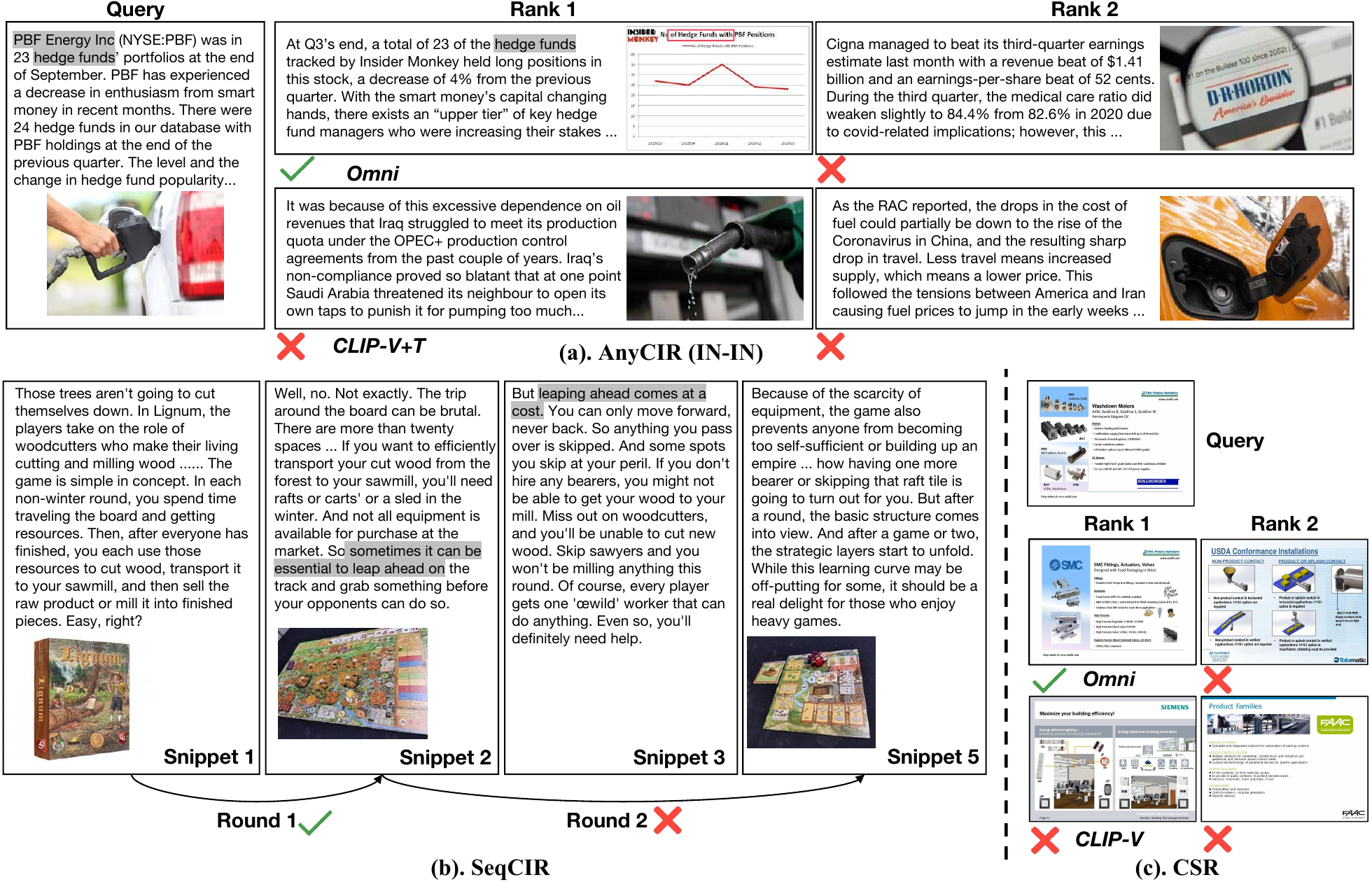}
\vspace{-15pt}
\caption{Visualization of retrieval results on AnyCIR, SeqCIR, and CSR benchmarks.}
\label{fig:vis}
\end{figure*}

\section{Ablation Study and Visualization}

\textbf{Effect of Model Initialization.}
As shown in Table~\ref{ab:model_init},
we observed that the CLIP initialization is important for \Mname.
Note that our training data only contains 5 million documents with around 17 million images, which is relatively small compared to WIT-400M. The scale-up experiments are left for future study due to the computational constraints and limited data scale.

\noindent\textbf{Importance of Image Rendering Positions.}
In Table~\ref{tab:ab_posgrid}, we ablate the effect of the image rendering position in girds as text content uses a fixed rendering order.
We rendered all the image content into the same grid positions for queries, while the candidates still use random positions.
The results indicate that \Mname~learns a robust representation against different rendered grid positions.

\noindent\textbf{Modality Masking and Text Masking Ratio Selection.}
In Table~\ref{tab:ab_mm}, we investigate the modality masking ratio of training data. It can be observed that modality masking is crucial for image-to-image retrieval ability learning.
In our setting, the best masking ratio is 40\% and the larger ratio will drop the performance.
Table~\ref{tab:ab_tm} reports the results of applying different text masking ratios during training. We find that randomly dropping sentences in the text can improve language understanding capacity. One possible reason is that the longer text has more redundant information.

\noindent\textbf{Non-Consecutive Pair Sampling.}
In Table~\ref{tab:ab_pospair}, we compare models using different ratios of one-hot consecutive pair for training.
Generally, more consecutive pairs achieve higher performance on the AnyCIR benchmark as these data are more aligned with AnyCIR tasks.
The one-hop consecutive pairs only slightly degrade the performance, which indicates model can learn useful representation from the non-consecutive snippets with a weaker connection.

\noindent\textbf{Retrieval Results Visualization.} 
As shown in Fig.~\ref{fig:vis}(a) \Mname~understands the loosely vision-language correspondence correctly while CLIP-V+T(feature averaging) is dominated by the image feature in AnyCIR IN-to-IN task.
In Fig.~\ref{fig:vis}(b), it can be observed that SeqCIR is a very challenging task as it requires the model to capture the precise connection between the consecutive snippets from omni-modality input.
Lastly, Fig.~\ref{fig:vis}(c) indicates that despite being trained on rendered data, \Mname~can effectively generalize to real-world complex layouts with different font sizes and styles.

\section{Conclusion and Limitations}
We introduce \Mname, a unified vision-centric framework that renders interleaved multimodal content directly in pixel space, enabling a simple yet effective contrastive learning approach without relying on modality-specific components. 
By leveraging the natural coherence in multimodal documents and applying snippet-level contrastive learning with masking-based augmentation, \Mname learns robust representations from loosely aligned, real-world multimodal web documents.
Our benchmarks validate that this vision-centric approach generalizes well across diverse retrieval scenarios and datasets.
We hope that \Mname~serves as a stepping stone for exploring multi-modal documents as valuable training data in the vision-language research community.

Although \Mname~can process any modality input using a single model from pixels, its efficiency and scalability are limited by its fixed input size.
Future work on designing a dynamic input strategy or new architecture could significantly enhance the performance and unlock more vision-centric applications for multi-modal web data understanding.

{\small
\bibliographystyle{unsrt}
\bibliography{main}
}
\appendix

\section{Broader Impact}
\label{sec:board_impact}

This work presents \Mname, a vision-centric contrastive learning framework designed to improve retrieval performance from complex, interleaved multimodal documents. By rendering both text and images into a unified pixel space and learning from consecutive web document snippets, \Mname~enables efficient and scalable retrieval ability across image, text, and image-text interleaved modality inputs.

\textbf{Positive Impacts}:
\Mname~has the potential to significantly enhance multimodal retrieval systems by supporting retrieval across heterogeneous content formats without complex preprocessing pipelines like OCR. This makes it particularly valuable for use in digital libraries, enterprise knowledge bases, and educational platforms, where documents often contain visual and textual information. The model’s simplicity and efficiency may also lower barriers to entry for organizations with limited computational resources.

\textbf{Negative Impacts}:
Improved retrieval capabilities may also carry risks. For example, the malicious user could exploit the retrieval ability to mine sensitive information from publicly available documents. Additionally, the reliance on pixel-based representations could reduce interpretability and obscure how retrieval decisions are made, potentially reinforcing hidden biases in the data.

\textbf{Mitigation Strategies}:
To address these concerns, we recommend incorporating content filtering, user access controls, and explainability features into any deployed retrieval systems based on \Mname. Ensuring that training data is diverse and ethically sourced is also critical to minimizing biases. Finally, robust monitoring procedures should be in place to detect and respond to misuse.

In summary, \Mname~offers a novel and practical solution for multimodal retrieval from complex document sources. However, responsible deployment and oversight are essential to mitigate potential risks and ensure its positive societal impact.

\section{More Implementation Details}
\label{sec:details}
\textbf{Data Pre-processing.}
Given a document, we chunked the document into several snippets in a sliding window strategy based on text sequence.
For MMC4~\citep{zhu2024multimodal}, the document text is stored in a list of sentences.
To create snippets, we merge consecutive sentences until their combined length reaches 1100 characters or less. 
Then we use the image-text assignment provided by MMC4 to assign each image to the corresponding snippet.
For OBELICS~\citep{laurenccon2024obelics}, we first split the text content based on the newline character and then use the same sliding window strategy to generate text snippets.
Differently, OBELICS organizes the documents as an image-text interleaved sequence, where the image position is extracted from the original HTML files.
In both AnyCIR and SeqCIR, we assign each image to the closest preceding text snippet, while images appearing at the beginning of the document are assigned to the first text snippet.

\noindent\textbf{Training Data Details.}
During training, to maintain optimal text length, we apply text masking augmentation only to snippets containing more than four sentences and exceeding 250 characters.
Empirically, we found that a maximum text length of 768 characters during training led to better performance. During testing, the model can handle up to 1,100 characters without any degradation in performance.
Therefore, we set the maximum training text length to 768 characters and 1,100 characters for the testing setting including AnyCIR, SeqCIR and MTEB~\citep{muennighoff2023mteb} benchmark.
After initialization from the CLIP pre-trained checkpoint, the positional embedding is randomly initiated for 448$\times$448 input size.
For each training batch, the data modalities are mixed from image, text, and image-text interleaved without specialized balance.

\section{Additional Experiment Analysis}
Table~\ref{ab:full_ab} presents the complete results of the AnyCIR benchmark (in total of 9 tasks) used in the ablation study, including model initialization, image rendering positions, modality masking ratio, text masking ratio and consecutive pair sampling. 
Moreover, we further provide the analysis of the text padding technique used in the M-BEIR~\citep{wei2023uniir} task.
Table~\ref{ab:text_pad} shows the ablation study on text padding to exceed a certain length by repeating it and its impact on the performance of the M-BEIR task. Note that the number of words of the query in this sub-task (image-text pair retrieval image-text pair) is often less than 10 words. The results suggest that the short text information might be surpassed in the image-text interleaved representation for OmniContrast in such cases.

\section{Visualization}
\begin{figure*}[h]
    \centering
    \includegraphics[width=\linewidth]{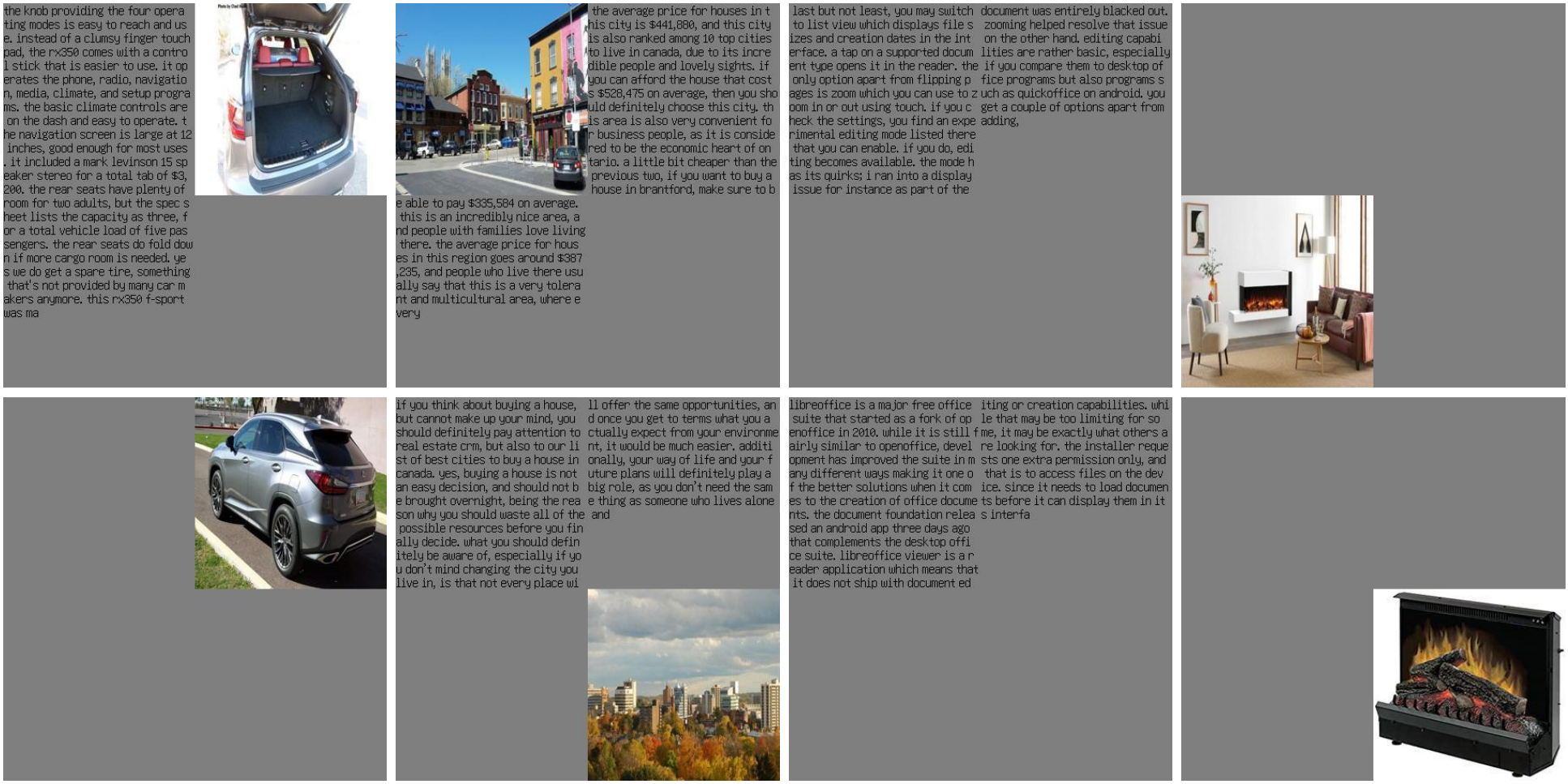}
    \caption{Rendered image-text snippets from a training batch. Each column represents the positive pairs.}
    \label{fig:vis_train}
\end{figure*}

\noindent\textbf{Training Data.} In Figure.~\ref{fig:vis_train}, we showcase some rendered snippet samples used for training from the MMC4 datasets in a batch.
We can observer that the model is trained for matching various target, i.e., interlevaed to image, interlevaed to interlevaed, text to text and image to image.
Note that the samples are rendered after applied with modality mask and text mask augmentations.

\begin{figure*}[h]
    \centering
    \includegraphics[width=\linewidth]{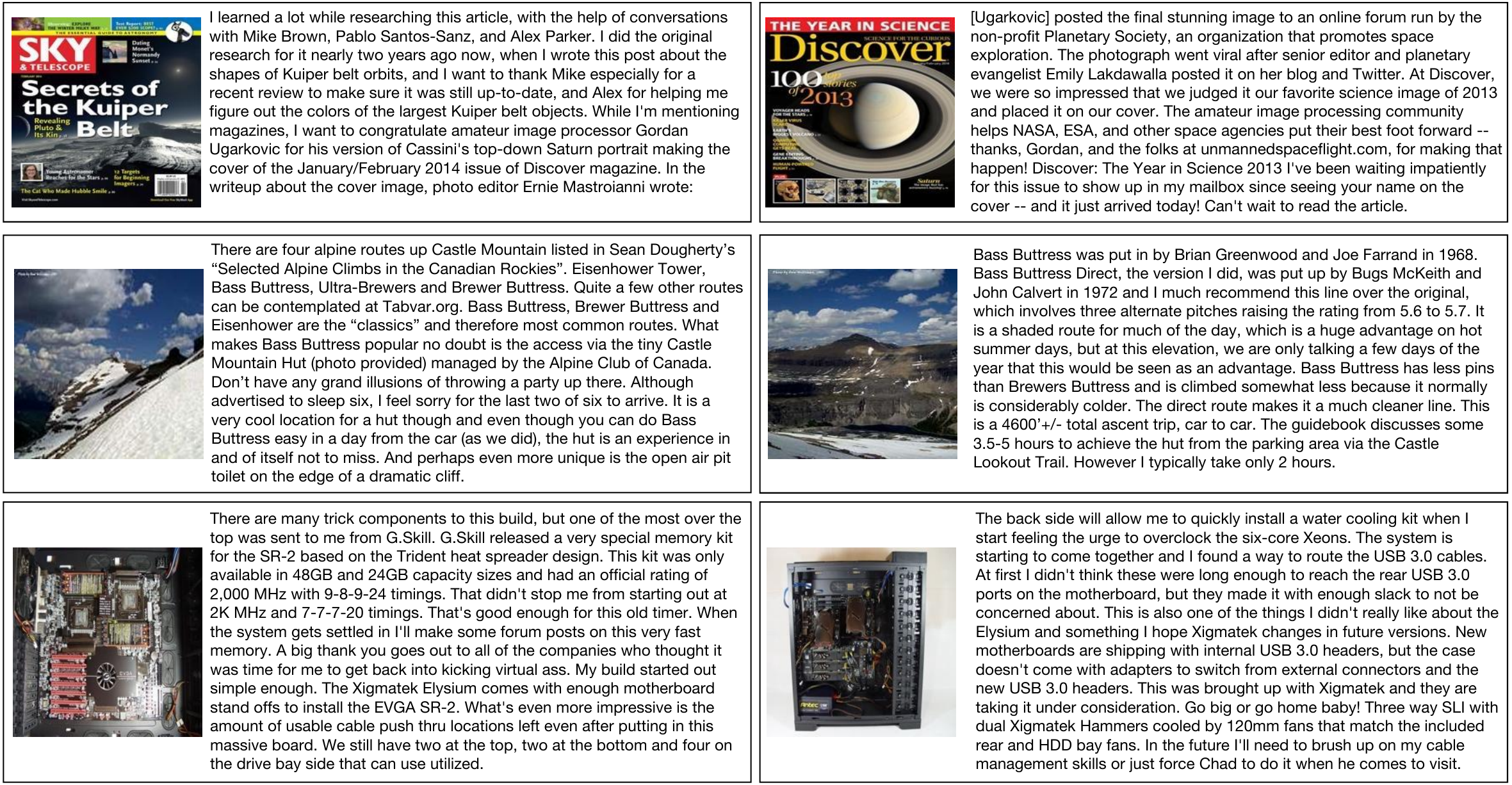}
    \caption{Visualization samples in AnyCIR benchmark. Each row represents the consecutive pairs.}
    \label{fig:vis_anycir}
\end{figure*}

\begin{figure*}[h]
    \centering
    \includegraphics[width=\linewidth]{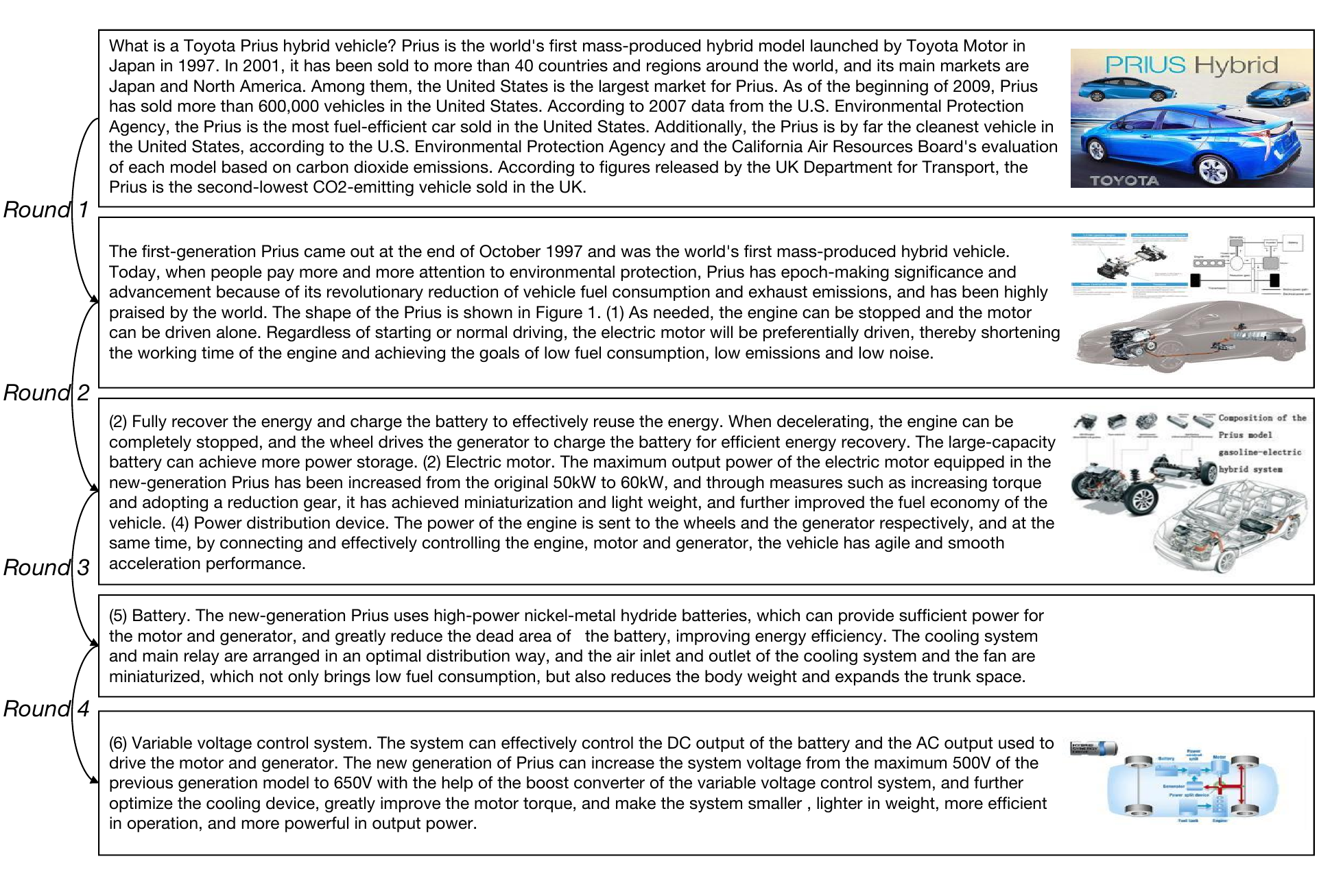}
    \caption{Visualization sample in SeqCIR benchmark.}
    \label{fig:vis_seqcir}
\end{figure*}

\begin{figure*}[h]
    \centering
    \includegraphics[width=\linewidth]{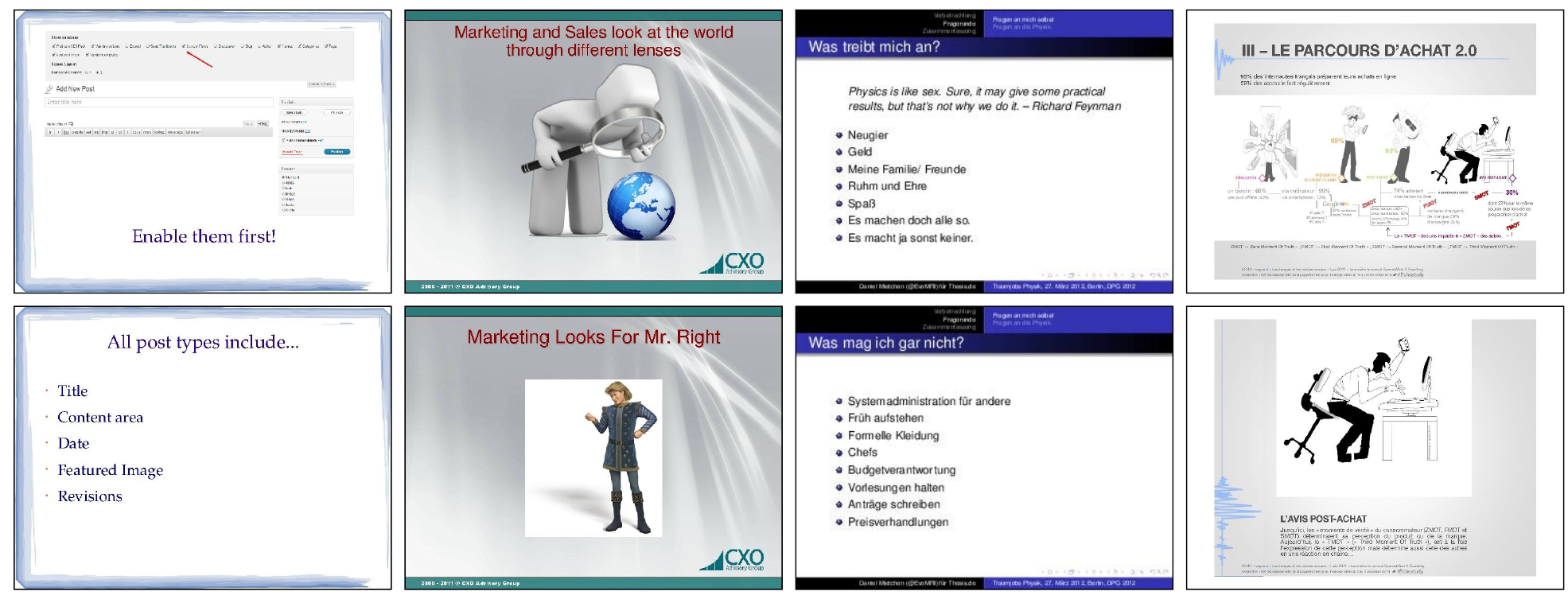}
    \caption{Visualization samples in CSR benchmark. Each column represents the consecutive pairs.}
    \label{fig:vis_csr}
\end{figure*}

\noindent\textbf{Benchmark Samples.} We further present more examples of our proposed consecutive information retrieval AnyCIR (Figure.~\ref{fig:vis_anycir}), SeqCIR (Figure.~\ref{fig:vis_seqcir}) and CSR benchmark (Figure.~\ref{fig:vis_csr}).
In Figure\ref{fig:vis_anycir}, we visualize the consecutive pairs sampled from a ddocumentn AnyCIR. It can be observed that the vision-language corresponding of these pairs is loose compared to the image-text caption data.
In Figure.~\ref{fig:vis_seqcir}, we visualize a full sequence of multi-round retrieval in SeqCIR, which is very challenging because the consecutive snippets within the same documents share high relevance.
In Figure.~\ref{fig:vis_csr}, we showcase some consecutive slides sampled from the slide desks~\citep{araujo2016large}.
Compare to the training data, the slide text are more short but with various layout and font size, which are out-of-domain data for OmniContrast.
Note that some slides share the same template, requiring models not only to focus on visual context but also on language content.
\begin{table*}[h]
\vspace{10pt}
\caption{Full results of ablation study in AnyCIR.}
\resizebox{\linewidth}{!}{%

\begin{tabular}{cc|ccccccccc|c}
\hline
\multicolumn{2}{c|}{Settings} & IN-IN & IN-Tx & IN-Im & Tx-IN & Tx-Tx & Tx-Im & Im-IN & Im-Tx & Im-Im & Overall \\ \hline
\multicolumn{1}{c|}{-} & IN-IN & 65.85& 64.26& 0.10& 63.84& 64.55& 0.05& 1.10& 0.19& 6.46& 29.60\\
\multicolumn{1}{c|}{Init $\checkmark$} & IN-IN & 76.56& 74.85& 0.40& 74.19& 74.81& 0.12& 2.58& 0.64& 8.95& 34.79\\
\multicolumn{1}{c|}{-} & Omni & 62.30& 59.29& 8.52& 59.11& 61.22& 1.47& 8.23& 1.49& 12.18& 30.42\\
\multicolumn{1}{c|}{Init $\checkmark$} & Omni & 78.27& 73.89& 22.10& 74.19& 74.32& 10.08& 22.00& 10.95& 19.50& 42.81\\ \hline
\multicolumn{1}{c|}{\multirow{4}{*}{\begin{tabular}[c]{@{}c@{}}Image\\ Rendering\\ Positions\end{tabular}}} & grid-0 & 78.17& 73.96& 22.15& 74.38& 74.32& 10.12& 22.07& 10.88& 19.53& 42.84\\
\multicolumn{1}{c|}{} & grid-1 & 78.26& 74.05& 22.07& 74.38& 74.32& 10.12& 22.18& 11.03& 19.50& 42.88\\
\multicolumn{1}{c|}{} & grid-2 & 78.31& 74.01& 22.00& 74.38& 74.32& 10.12& 22.01& 10.91& 19.51& 42.84\\
\multicolumn{1}{c|}{} & grid-3 & 78.18& 73.78& 22.04& 74.38& 74.32& 10.12& 22.18& 11.03& 19.43& 42.83\\ \hline
\multicolumn{1}{c|}{\multirow{6}{*}{\begin{tabular}[c]{@{}c@{}}Modality\\ Masking\\ Ratio\end{tabular}}} & 0.0 & 76.56& 74.85& 0.40& 74.19& 74.81& 0.12& 2.58& 0.64& 8.95& 34.79\\
\multicolumn{1}{c|}{} & 0.2 & 76.22& 71.47& 21.94& 71.44& 71.63& 10.67& 21.56& 11.25& 19.50& 41.74\\
\multicolumn{1}{c|}{} & 0.4 & 77.41& 72.06& 21.72& 72.74& 72.39& 9.71& 21.78& 10.72& 19.30& 41.98\\
\multicolumn{1}{c|}{} & 0.6 & 77.60& 73.35& 20.72& 72.90& 73.29& 9.02& 20.70& 9.47& 18.74& 41.75\\
\multicolumn{1}{c|}{} & 0.8 & 78.00& 74.32& 17.38& 73.93& 73.96& 6.89& 17.96& 7.69& 17.06& 40.80\\
\multicolumn{1}{c|}{} & 1.0 & 76.56& 74.49& 0.54& 74.07& 74.26& 0.26& 2.78& 0.65& 8.71& 34.70\\ \hline
\multicolumn{1}{c|}{\multirow{6}{*}{\begin{tabular}[c]{@{}c@{}}Text\\ Masking\\ Ratio\end{tabular}}} & 0.0 & 77.41& 72.06& 21.72& 72.74& 72.39& 9.71& 21.78& 10.72& 19.30& 41.98\\
\multicolumn{1}{c|}{} & 0.2 & 78.34& 73.96& 21.85& 74.25& 74.26& 10.16& 21.46& 10.89& 19.27& 42.71\\
\multicolumn{1}{c|}{} & 0.4 & 78.27& 73.89& 22.10& 74.19& 74.32& 10.08& 22.00& 10.95& 19.50& 42.81\\
\multicolumn{1}{c|}{} & 0.6 & 77.70& 73.44& 21.94& 73.42& 73.56& 10.11& 21.88& 10.77& 19.48& 42.48\\
\multicolumn{1}{c|}{} & 0.8 & 77.85& 73.20& 21.86& 73.20& 73.32& 10.11& 22.01& 10.64& 19.58& 42.42\\
\multicolumn{1}{c|}{} & 1.0 & 77.41& 72.38& 21.60& 72.66& 72.60& 9.67& 21.64& 10.61& 19.08& 41.96\\ \hline
\multicolumn{1}{c|}{\multirow{6}{*}{\begin{tabular}[c]{@{}c@{}}Consecutive\\ Pair\\ Sampling\end{tabular}}} & 0.0 & 78.27& 73.89& 22.10& 74.19& 74.32& 10.08& 22.00& 10.95& 19.50& 42.81\\
\multicolumn{1}{c|}{} & 0.1 & 78.04& 73.27& 21.88& 73.66& 73.53& 9.90& 21.96& 10.94& 19.74& 42.54\\
\multicolumn{1}{c|}{} & 0.2 & 77.94& 73.68& 21.87& 73.73& 73.68& 10.06& 21.76& 10.70& 19.29& 42.52\\
\multicolumn{1}{c|}{} & 0.3 & 78.13& 73.46& 21.46& 73.76& 73.65& 9.98& 21.51& 10.68& 19.31& 42.44\\
\multicolumn{1}{c|}{} & 0.4 & 78.05& 73.53& 21.27& 73.57& 73.41& 9.96& 21.48& 10.63& 19.55& 42.38\\
\multicolumn{1}{c|}{} & 0.5 & 77.95& 73.50& 21.29& 73.37& 73.54& 9.80& 21.59& 10.47& 19.29& 42.31\\ \hline
\end{tabular}
}
\label{ab:full_ab}
\vspace{10pt}
\end{table*}

\begin{table*}[h]
\caption{Ablation study of text padding length on M-BEIR benchmark.}
\centering
\begin{tabular}{c|c|ccccc}
\hline
\multirow{2}{*}{Task} & \multirow{2}{*}{Dataset} & \multicolumn{5}{c}{Text Padding Length} \\ \cline{3-7} 
 &  & - & 100 & 400 & 800 & 1000 \\ \hline
\multirow{2}{*}{$(q_i, q_t) \to (c_i, c_t)$} & oven\_task8 & 0.26 & 0.65 & 4.37 & 5.77 & 5.21 \\
 & infoseek\_task8 & 0.09 & 0.33 & 3.01 & 4.21 & 4.05 \\ \hline
\end{tabular}
\label{ab:text_pad}
\end{table*}


\clearpage

\end{document}